\documentclass[letterpaper, 10 pt, conference]{ieeeconf}  
\def\IEEEtitletopspace{0.25in}
\IEEEoverridecommandlockouts                              

\usepackage{adjustbox}
\usepackage{amssymb}
\usepackage{threeparttable}
\usepackage{subcaption}
\usepackage[font={footnotesize}]{caption}
\usepackage[table,dvipsnames]{xcolor}
\usepackage{multirow}
\usepackage{booktabs}
\usepackage{graphicx}
\usepackage{cite}
\usepackage{hyperref}

\newcommand{\refsec}[1]{Sec.~\ref{sec:#1}}

\newcommand{\reffig}[1]{Fig.~\ref{fig:#1}}

\newcommand{\reftbl}[1]{Table~\ref{tbl:#1}}

\DeclareGraphicsExtensions{.png,.pdf,.eps,.jpg}
\definecolor{not-sig}{gray}{0.75}

\def\figurePath{figs/}
\def\myfigurew#1#2#3{\begin{figure}[t!]\centering\includegraphics*[width = #3\linewidth]{\figurePath#1}\caption{#2}\label{fig:#1}\end{figure}}

\def\version#1{\textbf{v#1}}

\title{A Simulator Dataset to Support the Study of Impaired Driving}
\author{John Gideon$^*$, Kimimasa Tamura, Emily Sumner, Laporsha Dees, Patricio Reyes Gomez, \\ Bassamul Haq, Todd Rowell, Avinash Balachandran, Simon Stent, Guy Rosman
\thanks{All authors are with the Toyota Research Institute, USA. Email: \texttt{firstname.lastname@tri.global}. Any opinions, findings, and conclusions expressed in this material are those of the authors and do not necessarily reflect the views of TRI or any other Toyota entity.} 
\thanks{$^*$ Corresponding author.}}

\begin{document}

\maketitle

\thispagestyle{empty}
\pagestyle{empty}

\begin{abstract}
Despite recent advances in automated driving technology, impaired driving continues to incur a high cost to society. In this paper, we present a driving dataset designed to support the study of two common forms of driver impairment: alcohol intoxication and cognitive distraction. Our dataset spans 23.7 hours of simulated urban driving, with 52 human subjects under normal and impaired conditions, and includes both vehicle data (ground truth perception, vehicle pose, controls) and driver-facing data (gaze, audio, surveys). It supports analysis of changes in driver behavior due to alcohol intoxication (0.10\% blood alcohol content), two forms of cognitive distraction (audio n-back and sentence parsing tasks), and combinations thereof, as well as responses to a set of eight controlled road hazards, such as vehicle cut-ins. The dataset will be made available at \url{https://toyotaresearchinstitute.github.io/IDD/}.
\end{abstract}

\section{INTRODUCTION}
\label{sec:intro}

Human drivers are prone to decision-making failures while driving, due to a range of behavioral impairments. Such failures can have significant consequences for human life.
Two common sources of behavioral impairment for human drivers are alcohol intoxication and cognitive distraction.

\textbf{Alcohol intoxication}, even at moderate levels, is known to significantly increase the risk and severity of road traffic accidents. 
According to a recent report from the World Health Organization, about 20\% of fatally injured drivers in high-income countries have blood alcohol concentration (BAC) levels above the legal limit, while in low- and middle-income countries that number rises to between 33\% and 69\%~\cite{WHO2023}.

\textbf{Cognitive distraction (CD)} -- inattention to the task of driving as mental processes are diverted to other activities -- is far more prevalent in the driving population but poses relatively lower risk since it describes a transient mental state rather than lasting impairment.
Distracted driving is typically divided into three types -- visual (eyes off the road), manual (hands off the wheel), and cognitive (mind off the task)~\cite{AAA2013}. 
In the US, distracted driving contributes to around one-tenth of road traffic accidents~\cite{NHTSA2024}. 
While visual or manual distraction accounts for a lot of these incidents, cognitive distraction is the most difficult form of distraction to observe and measure~\cite{AAA2013}, making mitigation challenging.

\myfigurew{overview}{\textbf{Impaired Driving Dataset overview.} We captured data from 52 human drivers over 25 hours of urban driving in a driving simulator, including under \textit{alcohol-intoxicated} and \textit{cognitively-distracted} driving conditions and with a range of realistic \textit{driving hazards}. Our dataset supports various analyses including: the overlap between different types of impairment, how to distinguish one from the other, and their impact on behaviors such as visual attention and responses to road hazards.}{1.0}

The contributions of our paper are as follows:
\begin{itemize}
    \item We design and release the first publicly available driving simulator experiment that combines cognitive distraction (CD), alcohol intoxication, and road hazards, as illustrated in~\reffig{overview}.
    \item We run our experiment on 52 individuals, capturing a wide range of data including vehicle control during driving, eye gaze, ground truth scene state, ground truth impairment conditions, and various self-reported measures of driver state. 
    \item We validate our data by analyzing the relationship between several well-defined behavioral driving features and the various impairment conditions. 
\end{itemize}

\begin{table*}[t]
\centering
\caption{
\textbf{A summary of closely related datasets for studying simulated manual driving under impairment.} Our dataset is the first to explore the combination of alcohol intoxication, cognitive distraction and driving hazards, while providing coverage of driver gaze, controls and scene state. $^*$One hazard scenario involving unintended acceleration. $^{\dagger}$Gaze derived from RGB video. $^{\ddagger}$Cognitive workload (not distraction) is measured as a dependent variable.}
\begin{tabular}{l|ccc|ccc|ccc}
\toprule
& \multicolumn{3}{c|}{\textit{Variables of interest}} & \multicolumn{3}{c|}{\textit{Sensor data}} & \multicolumn{3}{c}{\textit{Dataset characteristics}} \\
Dataset	& Alcohol & Cog. Dist. & Hazards & Gaze track & Controls & Scene ground truth & Subjects & Dur. (hr) & Scenarios\\
\midrule
C42CN~\cite{Taamneh2017-sx} & - & \checkmark & \checkmark$^*$ & \checkmark & \checkmark & - & 68 & ~80 & Highway\\
CoCAtt~\cite{shen2022cocatt} & - & \checkmark & - & \checkmark & \checkmark & - & 11 & 12 & Countryside\\
Koch et al.~\cite{koch2023leveraging} & \checkmark & - & - & \checkmark & \checkmark & -	& 30 & 15 & Varied\\
Keshtkaran et al.~\cite{keshtkaran2024wacv} & \checkmark & - & - & -$^{\dagger}$ & - & - & 60 & 30 & Urban\\
CL-Drive~\cite{Angkan2024} & - & \checkmark$^{\ddagger}$ & - & \checkmark & - & - & 21 & 10 & Varied\\
Ours & \checkmark & \checkmark & \checkmark & \checkmark & \checkmark & \checkmark & 52 & 25 & Urban\\
\bottomrule
\end{tabular}
\label{tbl:datasets}
\end{table*}

By releasing our dataset to the community, we hope to encourage further work to develop systems that can quickly and accurately diagnose impaired driving in various forms, in the pursuit of improved road safety. The dataset will be made available at \url{https://toyotaresearchinstitute.github.io/IDD/}.

\section{RELATED WORK}
\label{sec:related}

\subsection{Alcohol intoxication and driving}
The physiological and behavioral changes exhibited by drivers under the influence of alcohol have been well studied and documented. These changes include a decreased capacity to process visual information~\cite{ziedman1980visualinfoproc}, 
an increased propensity to become distracted or not cope well with distractions~\cite{ahlstrom2023compensatory}
, a drop in vigilance~\cite{koelega1995vigilance}, and changes in visual behavior and eye movements linked to steering~\cite{tivesten2023influence}
. For a detailed review of these changes, please see~\cite{dong2024review}.

While alcohol impairment in the driving context is typically measured using field sobriety tests administered during roadside stops, various attempts have been made to harness vehicle information to predict driver intoxication in real-time.
Previous work has shown that it is possible to achieve performance comparable to the standardized field sobriety tests using eight minutes of driving observations~\cite{lee2010assessing}, although differences between drivers and roadway situations had a large influence on algorithm performance. 
Using further information from a driver-facing camera (to extract eye, gaze and head movement features), it has been shown that blood alcohol concentration can be predicted in a laboratory setting to reasonable accuracy~\cite{koch2023leveraging,keshtkaran2024wacv}.
In our study, we extend these prior efforts to more diverse driving conditions, incorporating realistic road hazards and cognitive distractions.  Additionally, we make the raw data available to support future analyses.

\subsection{Cognitive distraction and driving}
Cognitive distraction has been extensively explored in the driving literature \cite{Regan2011-ga,Yekhshatyan13}, in terms of underlying psychological and physiological phenomena \cite{hurts2011distracted, lee2014dynamics} and its impact on behavior and risk \cite{sussman1985driver,ascone2009examination,talbot2013exploring}.
It has been investigated for its interactions with other phenomena such as visual distraction~\cite{liang2010combining}, alcohol impairment~\cite{ahlstrom2023compensatory}, as well as its effect at different age groups~\cite{tanaka2020analysis}. 

Yet another thread of research involved possible approaches and interventions to address CD \cite{lee2004collision,lees2007influence,ahlstrom2013gaze,tanaka2020preliminary}. Several approaches have been used to detect both visual and CD in various conditions and dataset realism levels 
\cite{Kutila2007,yang2018analysis,sun2020research,Ahlstrom2021-ly,Misra2023}. However, these approaches were often explored within individual studies, making comparing them or reproducing them difficult, meriting a more comprehensive and unified dataset to serve as a benchmark within the research community

\subsection{Behavioral driving datasets}
Numerous on-road datasets exist for studying driver behaviors such as attention~\cite{ortega2020dmd}, gaze~\cite{kasahara22eccv}, in-cabin activities~\cite{yang2023aide}, or maneuver intent~\cite{jain2015car}. 
Studying the behaviors of drivers under impaired conditions presents a greater safety risk, hence the vast majority of datasets use driving simulators.

Several datasets that specifically address distraction have been released over the years \cite{Taamneh2017-sx,shen2022cocatt,Angkan2024}. We compare the most relevant datasets to ours in \reftbl{datasets}.
In general, such datasets are often limited to distraction alone, rather than including, e.g. interaction with alcohol impairment or curated interactions with specific road hazards. Moreover, many of them do not include vehicle controls or scene ground-truth information, which limits the features and patterns that can be observed and analyzed.

Our dataset is the first to mix the conditions of road risk and driver state. This makes it possible to study the joint and marginal effects of alcohol impairment and cognitive distraction during free driving and in response to realistic road hazards. We hope this dataset will alleviate some of the limitations of existing datasets towards reproducible and comprehensive research and evaluation of approaches for detecting cognitive distraction and intoxication, along with their interaction with both normal and hazardous driving conditions.

\section{METHOD}
\label{sec:study_design}

\begin{figure*}[t!]
    \centering
    \includegraphics[width=1.0\linewidth]{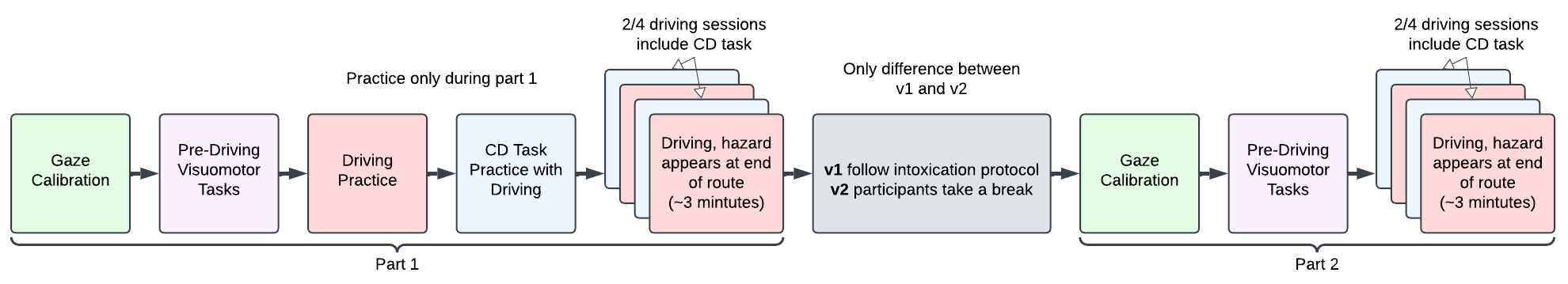}
    \caption{\textbf{Experimental flow overview,} consisting of two main parts separated by a break. The two experiment versions (v1 and v2) only differ in that v1 participants follow the intoxication protocol in \refsec{intoxication_protocol} and consume alcohol during the break.} \label{fig:experiment_flow}
\end{figure*}

\myfigurew{simulator}{\textbf{Driving simulator setup,} including steering wheel, pedals, tablet, and progress button to advance through the study. Driver-facing sensors included a Tobii Pro Spark eye tracker~\cite{tobii_user_guide}, headset microphone, and a webcam. Participants sat approximately 65 cm from the screen.}{0.83}

\begin{figure*}[t!]
    \centering
    \begin{subfigure}[t]{0.24\textwidth}
        \centering
        \includegraphics[height=0.7in]{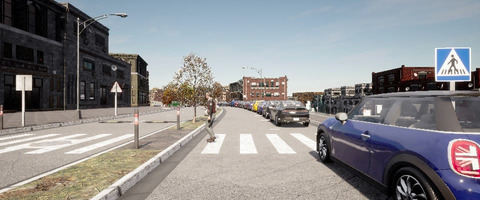}
        \caption{Pedestrian sudden crossing}
    \end{subfigure}%
    ~ 
    \begin{subfigure}[t]{0.24\textwidth}
        \centering
        \includegraphics[height=0.7in]{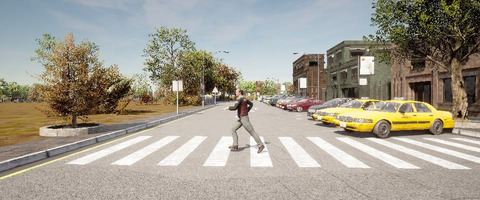}
        \caption{Pedestrian sudden crossing}
    \end{subfigure}%
    ~ 
    \begin{subfigure}[t]{0.24\textwidth}
        \centering
        \includegraphics[height=0.7in]{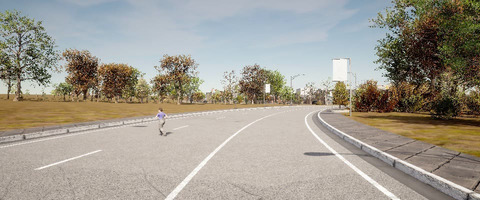}
        \caption{Child running out into road}
    \end{subfigure}%
    ~
    \begin{subfigure}[t]{0.24\textwidth}
        \centering
        \includegraphics[height=0.7in]{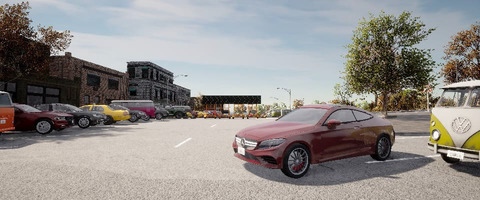}
        \caption{Car pop-out from parking spot}
    \end{subfigure}
    \begin{subfigure}[t]{0.24\textwidth}
        \centering
        \includegraphics[height=0.7in]{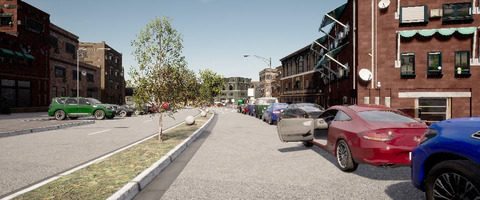}
        \caption{Car door opening}
    \end{subfigure}%
    ~
    \begin{subfigure}[t]{0.24\textwidth}
        \centering
        \includegraphics[height=0.7in]{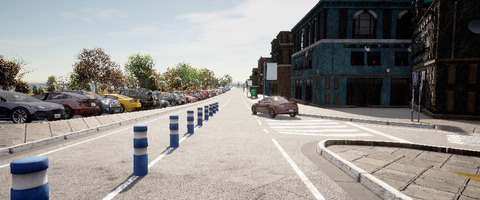}
        \caption{Car running stop sign merge}
    \end{subfigure}%
    ~ 
    \begin{subfigure}[t]{0.24\textwidth}
        \centering
        \includegraphics[height=0.7in]{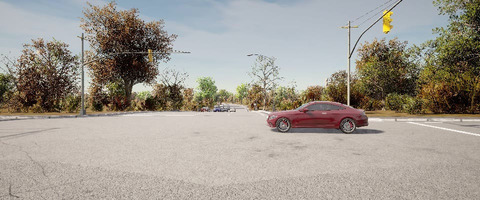}
        \caption{Car running red light}
    \end{subfigure}%
    ~
    \begin{subfigure}[t]{0.24\textwidth}
        \centering
        \includegraphics[height=0.7in]{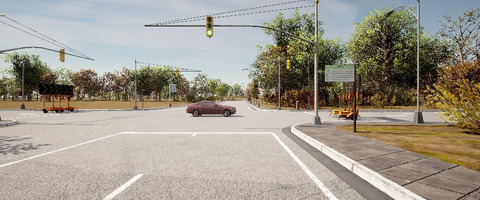}
        \caption{Car running red light}
    \end{subfigure}
    \caption{\textbf{The dataset contains a wide variety of aligned hazardous events,} allowing for further investigation on the impact of impairment on driving risk.
    \label{fig:hazards}}
\end{figure*}

\subsection{Experimental procedure}
To understand the effects of CD and alcohol intoxication, we executed two versions of the study: impaired driving with alcohol intoxication and CD (version 1, \version{1}) and driving only with CD (version 2, \version{2}). The overall experimental flow is visualized in \reffig{experiment_flow}.

\version{1} participants, who went through the intoxication procedure, were full-time Toyota Research Institute employees, while \version{2} participants were externally recruited through User Interviews\cite{user_interviews}. All participants signed a consent form prior to participation.  The experimental procedure was reviewed and approved by the WCG IRB (Protocol \#20241945). All personally identifiable information is stored on HIPAA-compliant machines to ensure participant privacy.

The study consisted of five primary components: 1) gaze calibration, 2) pre-driving visuomotor tasks, 3) driving practice, 4) CD task practice, and 5) driving with hazards. Participants were also given questionnaires at different time points in the experiment. Each participant went through these components twice. After completing it the first time, \version{1} participants followed the intoxication procedure while \version{2} participants took a break.  All study activity, with the exception of the intoxication procedure, took place in the driving simulator (\reffig{simulator}). 

Prior to driving, participants' gaze tracking was calibrated using the Tobii Spark Calibration~\cite{tobii_user_guide}. Participants then spent 10 minutes performing some simple pre-driving visuomotor tasks as part of another experiment~\cite{stent2025iv}. Next, participants began the driving portion of the study, which was performed in the CARLA~\cite{dosovitskiy2017carla} simulator's Town 15 map. Participants drove through a set route around the town for two minutes. Afterward, participants were introduced to each of the CD tasks: the 1-back \cite{he2019high} and sentence comprehension task \cite{baumann2009comprehension}. Participants would practice each task twice, once while they were not driving and once while they were driving. Participants only practiced the CD tasks during the first part of the experiment. 

Participants then drove a series of four scenarios of a set distance lasting approximately three minutes. Each of these ended with a hazard (depicted in \reffig{hazards}) meant to induce a reaction in the driver. These hazards were designed to generate interesting behaviors and test different aspects of driving, such as reaction time and lateral scene awareness. We use scripted hazards to ensure reproducibility and comparability between all participants. The hazard logic was designed to allow flexibility to different driver speeds and to adjust the respective pedestrian or other vehicle speed accordingly. Out of these four scenarios per block, two were carried out with no cognitive distraction and two were carried out with the added 1-back and sentence comprehension tasks (one each). This resulted in four routes driven per block, totaling eight routes driven and eight hazards across both parts per participant.

We employed four different orders in which participants experienced scenarios to ensure that each hazard had an even balance of baseline, CD tasks, and intoxication. After participants completed the first four scenarios, they began the intoxication portion (\version{1}) or took a break (\version{2}).

For each participant, the dataset includes several questionnaires from different time points in the experiment. We collected short-form PANAS~\cite{thompson2007development} questionnaires before, and after each part, as well as NASA-TLX~\cite{cao2009nasa} and Karolinska Sleepiness Scale (KSS)~\cite{aakerstedt1990subjective} after each hazard.

\subsection{Intoxication Procedure} \label{sec:intoxication_protocol}
\myfigurew{BAC_Levels}{\textbf{Participant BAC levels} at the start and end of the intoxicated portion.}{0.65}

\version{1} participants were instructed to not eat or drink anything for at least three hours before their study start time. Before beginning the experiment, they provided their weight and confirmed that they had a blood alcohol level (BAC) of 0 through a breathalyzer. The participant's weight and gender were used to determine how many grams of alcohol they needed to reach a BAC of 0.1. This threshold was selected, as prior work has shown an exponential increase in risk after a BAC of 0.1 \cite{blomberg2009long}. Participant BAC levels were monitored throughout the study, including before and after the intoxicated portion, as seen in~\reffig{BAC_Levels}.

Upon completion of the first part of the study, the proctor measured the correct amount of alcohol using a kitchen scale. Participants chose between tequila, vodka, or whiskey and were provided 1 oz of juice that could be used as a chaser or mixed in the drink. Participants had 10 minutes to consume all or as much of the alcohol as they could. The proctor assured participants that they were not required to drink all of the alcohol and could end the study at any point. Afterwards, BAC levels were checked and recorded every 10 minutes.  Once a participant had a BAC over 0.1 and their BAC was decreasing, participants went back to the simulator room to complete the second half of the study. If a participant had a BAC under 0.08, the wellness proctor would ask if the participant would be comfortable consuming an additional 20g of alcohol. If a participant said yes then the additional alcohol would be provided and noted.

Once participants completed the impaired driving portion of the study, the proctor escorted participants to a recovery space, where they remained until reaching a BAC level appropriate for their return to work or home.

\section{DATASET}
\label{sec:dataset}

\subsection{Participants}
In \version{1} (n = 20), ages ranged from between 23-45 with the average age being 32.2 (SD 6.4). Thirteen participants identified as male, six identified as female, and one identified as non-binary. Ethnically, most identified as White (8) or Asian (7), leaving those who identified as Black/African American (1), Native Hawaiian or Other Pacific Islander (1), and mixed-raced (2) in the minority (one participant did not report their race/ethnicity and another did not report any of their demographics). There were initially 22 participants, 1 participant was excluded due to participating in a pilot with slightly different procedures. The other participant dropped out after completing the first section. 

In \version{2} (n = 32) ages ranged from 23 to 65, with a mean of 37.1 (SD 10.3). This portion included 18 participants who identified as male, 13 as female, and 1 as non-binary. A total of 9 participants identified ethnically as White, 9 as Black/African American, 7 as Asian, and 7 as Hispanic/latino. Of the initial 38 \version{2} participants, 6 were excluded for dropping out of the study due to motion sickness.

\begin{table}[t]
\centering
\caption{
\textbf{Amount of data in each driver state.}
}
\begin{tabular}{cc|cccc}
\toprule
\multicolumn{2}{c|}{\textit{Driver State}} & & & Distance & Valid \\
Alcohol & CD & Trips & Hours & in km & Hazards \\
\midrule
No & No & 168 & 9.9 & 342 & 160 \\
No & Yes & 168 & 9.7 & 336 & 158 \\
Yes & No & 40 & 2.1 & 81 & 40 \\
Yes & Yes & 40 & 2.1 & 81 & 37 \\
\midrule
\multicolumn{2}{c|}{Total} & 416 & 23.7 & 839 & 395 \\
\bottomrule
\end{tabular}

\label{tbl:data_amounts}
\end{table}
\myfigurew{HazardMontageNew}{\textbf{A montage of screenshots from a subset of participant experiences of the ``pedestrian sudden crossing'' hazard.}
Hazards were triggered depending on each participant's speed leading up to the hazard location. Trigger logic was tuned to provide the most similar experience within normal speed bounds, allowing for direct comparisons of hazard responses across participants. We labeled 5\% (22/416) of cases where the hazard did not deploy correctly, due to the participant driving off-route or excessively speeding, to remove them from any analyses.}{1.0}

\subsection{Dataset Structure}
We collected an overall dataset of $23.7$ hours driving, for a total of $839$ km, and a total of $395$ hazard instances under various conditions of impaired/normal driving -- see \reftbl{data_amounts} for the full breakdown of driving data collected and \reffig{HazardMontageNew} to see alignment within a single hazard.

In terms of scene information, forward-facing virtual RGB, depth, and semantic segmentation camera from CARLA's virtual sensors (sampled at 10 Hz and a $960 \times 400$ resolution) were recorded. 
CARLA ego state and controls were collected at 10 Hz, as well as and the other vehicle tracks. The other vehicle tracks were collected at 10 Hz from the CARLA state, with some impact (70/416 trips) due to an unknown simulator traffic spawning issue, which we annotated as unavailable.
These state and scene logs were combined with gaze information --- for gaze logging, Tobii Spark Pro~\cite{tobii_user_guide} measurements were recorded at 60 Hz, and contained gaze vectors for each eye, gaze point on screen, and pupil diameter.

\begin{table}[t]
\centering
\caption{
\textbf{A breakdown by driver state of ego-vehicle collisions (crashes) incurred.} Middle: only counting collisions during one of the eight designed road hazards. Right: counting all collisions. Alcohol more than doubles the rate of collisions, while cognitive distraction only increases it marginally.}
\begin{tabular}{cc|cc|cc}
\toprule
\multicolumn{2}{c|}{\textit{Driver State}} & \multicolumn{2}{c|}{\textit{Designed Hazard}} & \multicolumn{2}{c}{\textit{All Crashes}}\\
Alcohol & CD & Crash & Crash/Hazard & Crash & Crash/Trip \\
\midrule
No & No & 10 & 0.06 & 48 & 0.29 \\
No & Yes & 10 & 0.06 & 52 & 0.31  \\
Yes & No & 5 & 0.13 & 26 & 0.65 \\
Yes & Yes & 8 & 0.22 & 26 & 0.65 \\
\midrule
\multicolumn{2}{c|}{Total} & 33 & 0.08 & 152 & 0.37 \\
\bottomrule
\end{tabular}

\label{tbl:crash_risk}
\end{table}

\subsection{Collisions}
Collisions with other vehicles and pedestrians occurred throughout the collection of the study - both due to planned hazard interactions and emergent encounters mostly due to the dynamics of the lead vehicle. \reftbl{crash_risk} shows the distribution of these crashes across the various driver states. While there was little effect seen for sober CD, intoxicated participants had a substantially higher collision rate of 0.13 versus 0.06 in designed hazards and 0.65 versus 0.29 throughout all driving.

\subsection{Simulation Anomalies} \label{sec:issues}
We relied on randomly generated traffic patterns during the part of the experiment preceding the hazard both to allow for a wider variety of interactions and to make larger numbers of vehicles feasible to implement. However, this occasionally produced anomalous behaviors - such as traffic coming to a complete standstill. Because of this, once the data collection was complete, we visually inspected the entire dataset to generate temporal annotations for anomalous traffic events, in case it becomes necessary to remove them in future research.

\textbf{Mistriggered hazards:} Due to likely physics issues or participants driving too quickly, some of the hazards did not trigger as planned (as seen in \reftbl{data_amounts}) and are annotated.

\textbf{Despawning vehicles:} In initial piloting it was found that participants could get into traffic deadlocks. To mitigate this, we despawn any vehicle that remains still for more than five seconds. This sometimes results in the disappearance of vehicles within view of the driver, especially near intersections. A list of these despawns and causes are included with the dataset.

\textbf{Vehicle Displacement:} An unknown simulation bug sometimes caused a stationary car to appear slightly off the road, affecting 1.5\% of captured driving data. The start and end times of these occurrences are annotated.

\textbf{Participant off route:} A total of eight participants drove substantially off the specified route during one of their scenarios. Of these, six returned to complete the route. The times of these off-route portions are annotated in the dataset.

\textbf{Other vehicles driving through barriers:} We used a set of traffic barriers to define the route to follow. However, since we wanted the other vehicles to be able to drive other routes, we allowed for them to drive through these barriers.

\section{DATASET VALIDATION}
\label{sec:validation}

\begin{table*}[t]\centering
  \begin{tabular}{l|rr|rr|rr}
\toprule
 & \multicolumn{2}{c}{Baseline vs Intox.} & \multicolumn{2}{c}{Baseline vs CD} & \multicolumn{2}{c}{Baseline vs Both} \\
Feature & Diff. & p-value & Diff. & p-value & Diff. & p-value \\
\midrule

Longitudinal Speed (m/s) & \textcolor{not-sig}{0.796} & \textcolor{not-sig}{0.1231} & \textcolor{not-sig}{-0.028} & \textcolor{not-sig}{0.9202} & 1.180 & 0.0484 \\
Longitudinal Acceleration (m/s\textsuperscript{2}) & 0.117 & 0.0136 & -0.064 & 0.0080 & \textcolor{not-sig}{0.069} & \textcolor{not-sig}{0.1893} \\
Braking Acceleration (m/s\textsuperscript{2}) & 0.134 & 0.0032 & \textcolor{not-sig}{-0.020} & \textcolor{not-sig}{0.5418} & 0.117 & 0.0441 \\

\rule{0pt}{3ex}Steering Reversals at 0.5 Degrees (\#/min) & 3.400 & 0.0012 & \textcolor{not-sig}{0.529} & \textcolor{not-sig}{0.3198} & 3.633 & 0.0066 \\
Steering Reversals at 2.5 Degrees (\#/min) & 1.233 & 0.0187 & \textcolor{not-sig}{-0.109} & \textcolor{not-sig}{0.3439} & \textcolor{not-sig}{0.817} & \textcolor{not-sig}{0.0826} \\

\rule{0pt}{3ex}Gaze Yaw Standard Dev. (Degrees) & -0.989 & 0.0049 & -0.514 & 0.0014 & -1.042 & 0.0010 \\

\rule{0pt}{3ex}Fixations (\#/min) & -41.517 & 0.0000 & -21.324 & 0.0000 & -55.883 & 0.0001 \\
Saccades (\#/min) & -68.683 & 0.0000 & -19.606 & 0.0011 & -78.167 & 0.0000 \\

\rule{0pt}{3ex}Pupil Diameter (mm) & \textcolor{not-sig}{-0.007} & \textcolor{not-sig}{1.0000} & 0.073 & 0.0000 & \textcolor{not-sig}{0.116} & \textcolor{not-sig}{0.0696} \\
\bottomrule
  \end{tabular}
  \caption{The results of our dataset validation. We consider four different pairwise comparisons and report their distribution mean differences and the p-value calculated using a Wilcoxon signed-rank test. We gray out any result with a p-value greater than 0.05 as not significant.} \label{tbl:results}
\end{table*}

We next demonstrate the ecological validity of our dataset for the study of alcohol intoxication and cognitively distracted driving by analyzing sets of well-known behavioral features established in prior work. We consider both control and gaze input modalities. When considering the gaze modality, we take the average of the left and right eyes.

For each feature, we first calculate its value in all scenarios. For consistency, we only consider the scenario data from 120 seconds before the end of the recording and 30 seconds before (to only focus on driving without hazards). We then consider three different driver state comparisons for analysis - (i) baseline (instances without either form of impairment) versus intoxication, (ii) baseline versus CD, and (iii) baseline versus both. For each comparison, we group the scenarios based on the conditions of the experiment and take the mean of the feature. We then perform a Wilcoxon signed-rank test to determine if there is a significant difference between the features in the two conditions. Because only a subset of participants were intoxicated during the experiment, only 20 participants are used for the (i) and (iii) comparisons, while all 52 are used for (ii). We report both the p-value of this test and the difference of the means in \reftbl{results}.

\subsection{Speed and Acceleration}
The disinhibition caused by alcohol has been found to correlate with higher speeds and accelerations when driving \cite{fillmore2008acute}. However, there is not yet consensus in the field for the impact of CD on driving performance \cite{kountouriotis2016identifying}. In order to calculate vehicle dynamics at a particular instant, we consider the 5-second window centered on the current time. We then fit a cubic spline weighted with a Hann window and calculate the longitudinal speed and acceleration. Our results show mean speed in meters per second (m/s) and mean accelerations in meters per second squared (m/s\textsuperscript{2}).

Our analysis found that an increase in speed was only significant when participants were impacted by both intoxication and CD. Both types of acceleration were found to be significantly higher in intoxication, as in \cite{fillmore2008acute}, regardless of the presence of CD. CD had a mixed impact on acceleration -- significantly lowering forward acceleration when compared with baseline. This interfered with the effect of intoxication on acceleration and resulted in the increase no longer being significant when both states occurred simultaneously. 

\subsection{Steering Reversals}
The steering reversal rate (SRR) aims to capture the amount of steering corrections needed by a driver to follow a route. Kountouriotis et al. noted an increase in SRR when drivers experienced different forms of CD \cite{kountouriotis2016identifying}, while Li et al. found a similar increase during intoxication \cite{li2019effects}. We follow the work of Markkula et al. to calculate SRR \cite{markkula2006steering} and select thresholds of 0.5 and 2.5 degrees, as in \cite{kountouriotis2016identifying}. We report the counts of SRR per minute as the feature.

Intoxication resulted in a significant increase in both types of corrections versus baseline, as seen in \cite{li2019effects}. However, CD did not have a significant difference and resulted in mixed effects when considering with intoxication jointly. The minor steering reversals at 0.5 degrees were found to occur at a slightly higher rate when jointly modeled than when intoxicated alone, potentially indicating some effect, as in \cite{kountouriotis2016identifying}.

\subsection{Gaze Yaw Standard Deviation}
Prior work has noted that gaze becomes more concentrated on the road when cognitively loaded \cite{kountouriotis2016identifying} or intoxicated \cite{tivesten2023influence}. One method to measure this is to take the standard deviation of the yaw of the gaze vector, as in \cite{kountouriotis2016identifying}, which should decrease as the gaze becomes more concentrated. We report gaze yaw standard deviation in degrees.

We noted a significant decrease in gaze yaw standard deviation (increased concentration) in both intoxicated and CD experiments, as noted in \cite{kountouriotis2016identifying} and \cite{tivesten2023influence}. This compounded to an even stronger concentration when jointly modeled.

\subsection{Fixation and Saccade Counts}
Fixations are the periods when eye position remains relatively stable, allowing for detailed visual processing of an object or location, while saccades are the quick movements in between. Prior work has found that the occurrence of each of these decreases with both intoxication \cite{silva2017effects} and CD \cite{walter2021cognitive}. We use PyGaze \cite{dalmaijer2014pygaze} to extract the instances of fixations and saccades and obtain the counts per minute as a feature.

Both fixations and saccades significantly decreased in both driver states, as noted in prior works \cite{silva2017effects, walter2021cognitive}. The effect further increases when jointly modeled.

\subsection{Pupil Diameter}
Prior work has found that pupil size increases while engaged in a cognitive task \cite{walter2021cognitive}, while it decreases while intoxicated \cite{jolkovsky2022impact}. We calculate the mean pupil diameter across each scenario and report the amount in millimeters (mm).

Our results only show a significant increase in pupil size when cognitively distracted (as in \cite{walter2021cognitive}), with no effect seen for intoxication. Even though the effect size increases when under both intoxication and CD, the difference is not significant. This could indicate that more participants are needed to validate the joint effects of intoxication and CD, or point to the multifaceted nature of pupil response time signals \cite{mathot2018pupillometry,zandi2021deep}, which can be explored in our dataset. This could include variability due to screen brightness and participant fatigue, which may need to be explicitly modeled for a more thorough future analysis.

\section{LIMITATIONS}
\label{sec:limitations}
While the dataset contains some simulation anomalies, as noted in \refsec{issues}, we have done our best to annotate these to support data filtering at varying levels of strictness. In addition to these, there are a few broader limitations of the data collection methodology. While the dataset is comparable in participant count and duration to similar driver impairment-focused datasets~\cite{Taamneh2017-sx,shen2022cocatt,koch2023leveraging,keshtkaran2024wacv,Angkan2024}, it is relatively small when compared with many (particularly autonomous) driving datasets.  Regardless, we aim to continue to increase the size and diversity of this dataset in future work. One area for improvement will be to incorporate more realistic cognitive tasks. While 1-back and sentence compression tasks are typical CD baselines, they do not necessarily reflect typical in-vehicle behavior. Simulated driving datasets are also ultimately limited by the realism of the scene and the naturalness of the behavior. But due to safety concerns, it is difficult to collect a controlled experiment with intoxication in vehicle, especially one containing risky scenarios~\cite{koch2023leveraging,keshtkaran2024wacv}.

\section{CONCLUSIONS}
\label{sec:discussion}

We have created the first combined driving dataset containing annotated examples of multiple forms of driver impairment. We showed that our dataset has reasonable ecological validity by examining how various gaze and control-based features change under different types of impairment. We found that gaze features are generally useful for detecting both types of driver impairment, while control features are more effective for intoxication and pupil diameter is more effective for CD. These features tend to correlate to what has been found in prior work, validating that the dataset contains expected phenomena of intoxication and CD. However, when considering both types of impairment simultaneously, these features sometimes have constructive or destructive interference with one another. This shows the benefit of jointly collected data, as we are able to start disentangling the effects from one another. Furthermore, our collision statistics (\reftbl{crash_risk}) indicate that a better understanding of driver impairment could result in improved design of interventions, which could be directly tested using this dataset.

Our dataset supports a variety of research areas including: (i) online / multi-task impairment prediction, (ii) driver hazard response prediction, (iii) speed profile prediction, and (iv) gaze prediction. In particular, each area would be of interest when conditioned on known impairment and driver gaze. We view our dataset as a complement to real-world data that allows for deeper exploration of the intersection between impairment and responses to hazardous events in a way never before publicly available.

\bibliographystyle{IEEEtran}
\bibliography{references}

\end{document}